\documentclass[letterpaper,times]{IONconf}
\usepackage[pdftex]{graphicx}
\usepackage{url}

\usepackage{natbib}

\usepackage[hidelinks]{hyperref}

\usepackage{mathtools}
\usepackage{amssymb}
\usepackage{bm}
\usepackage{bbm}
\usepackage{caption}
\usepackage{subcaption}
\usepackage{xcolor}
\usepackage{outlines}
\usepackage{siunitx}

\title{Neural Radiance Maps for Extraterrestrial Navigation and Path Planning}

\author{
    Adam~Dai, Shubh~Gupta, and Grace~Gao, \textit{Stanford~University}
}

\begin{document}
\newcommand{\todo}[1]{{\color{Red}TODO: #1}}

\newcommand{\red}[1]{{\color{red}{#1}}}

\newcommand{\regtext}[1]{\mathrm{\textnormal{#1}}}

\newcommand{\x}{\mathbf{x}}
\newcommand{\w}{\mathbf{w}}
\newcommand{\view}{\mathbf{v}}
\newcommand{\f}{\mathbf{f}}
\newcommand{\rgb}{\mathbf{c}}

\newcommand{\R}{\ensuremath{\mathbb{R}}}

\newcommand{\var}{\mathrm{var}}
\newcommand{\mbf}[1]{{\mathbf{#1}}}
\newcommand{\transpose}{^\mathsf{T}}
\newcommand{\norm}[1]{\left\Vert#1\right\Vert}

\newcommand{\start}{_\regtext{start}}
\newcommand{\goal}{_\regtext{goal}}
\newcommand{\local}{_\regtext{local}}
\newcommand{\globl}{_\regtext{global}}
\newcommand{\thresh}{_\regtext{thresh}}
\newcommand{\rays}{_\regtext{rays}}

\maketitle

\section*{biography}


\biography{Adam~Dai}{is a Ph.D. candidate in the Department of Electrical Engineering at Stanford University. He graduated from Caltech with a B.S. in Electrical Engineering in 2019. His research focuses on integrating mapping and planning for safe navigation and autonomy.}

\biography{Shubh~Gupta}{is a Ph.D. candidate in the Department of Electrical Engineering at Stanford University. He graduated from the Indian Institute of Technology, Kanpur with a B.Tech in Electrical Engineering and a minor in Computer Architecture. He obtained his M.S. in Electrical Engineering from Stanford University. His research focuses on multi-sensor state estimation, perception, and uncertainty quantification for navigation and autonomy.
}

\biography{Grace Gao}{is an assistant professor in the Department of Aeronautics and Astronautics at Stanford University.
Before joining Stanford University, she was an assistant professor at the University of Illinois at Urbana-Champaign. She obtained her Ph.D. degree at Stanford University. Her research is on robust and secure positioning, navigation, and timing with applications to manned and unmanned aerial vehicles, autonomous driving cars, as well as space robotics.}


\section*{Abstract} 

Autonomous vehicles such as the Mars rovers currently lead the vanguard of surface exploration on extraterrestrial planets and moons.
In order to accelerate the pace of exploration and science objectives, it is critical to plan safe and efficient paths for these vehicles.
However, current rover autonomy is limited by a lack of global maps which can be easily constructed and stored for onboard re-planning.
Recently, Neural Radiance Fields (NeRFs) have been introduced as a detailed 3D scene representation which can be trained from sparse 2D images and efficiently stored.
We propose to use NeRFs to construct maps for online use in autonomous navigation, and present a planning framework which leverages the NeRF map to integrate local and global information.
Our approach interpolates local cost observations across global regions using kernel ridge regression over terrain features extracted from the NeRF map, allowing the rover to re-route itself around untraversable areas discovered during online operation.
We validate our approach in high-fidelity simulation and demonstrate lower cost and higher percentage success rate path planning compared to various baselines.

\section{Introduction}
\label{sec:intro}

In recent years, there have been renewed efforts in space exploration, as evidenced by the Curiosity (2012) and Perseverance (2020) Mars rover missions, as well as the planned Artemis missions to the Moon. 
Developing autonomous navigation capabilities is crucial for advancing these planetary exploration efforts.
Given that the communication time from Earth to Mars can take up to 20 minutes, real-time teleoperation of Mars rovers is infeasible. 
Currently, much of the Mars rover path planning has been done manually in advance with the assistance of human operators~\citep{sherwood2001integrated}. 
This human-in-the-loop approach is difficult to scale for larger missions or more complex environments. 
As such, there exists an opportunity for autonomous systems to facilitate more robust and efficient exploration.

A major challenge in autonomous planetary exploration is planning paths for a rover to traverse the terrain while accomplishing the mission objectives, avoiding hazards, and minimizing the wear and tear to hardware. 
Current approaches rely heavily on onboard cameras to determine traversability and generate cost maps for path planning. 
For example, NASA has developed an autonomous navigation (AutoNav) system for Mars rovers which processes onboard imagery to determine traversability and generate a corresponding cost map~\citep{carsten2007global}. 
However, this approach can result in suboptimal path planning due to the limited range of onboard cameras and their inability to detect occluded hazards. 
Additionally, onboard cameras may struggle to identify non-geometric hazards such as loose sand. 
Thus, there is a need to incorporate global information about the terrain to improve path planning and navigation capabilities. 

Global information for path planning is available through multiple resources, including data harvested from prior missions, satellite imagery databases, topography maps, and drone imaging. 
The Mars 2020 mission serves as a recent case in point, wherein the Mars Reconnaissance Orbiter (MRO) captured images that were used in optimal route planning for landing site selection~\citep{ono2016data}. 
In a similar vein, \textit{Ingenuity} helicopter's images were utilized to identify suitable locations ahead of the mission~\citep{tzanetos2022ingenuity}. 
The planning process driven by these global sources of information employs a wide range of techniques like terrain classification, rock detection, and stereo processing to create a mobility model or cost function for the terrain. 
This model encodes traversibility aspects, such as estimated driving speed for the rover, and integrates them with a sequential path planning algorithm to compute the most efficient path for visiting multiple regions of interest~(ROIs). 

\begin{figure}[!ht]
     \centering
     \begin{subfigure}[b]{0.4\textwidth}
         \centering
         \includegraphics[width=\textwidth]{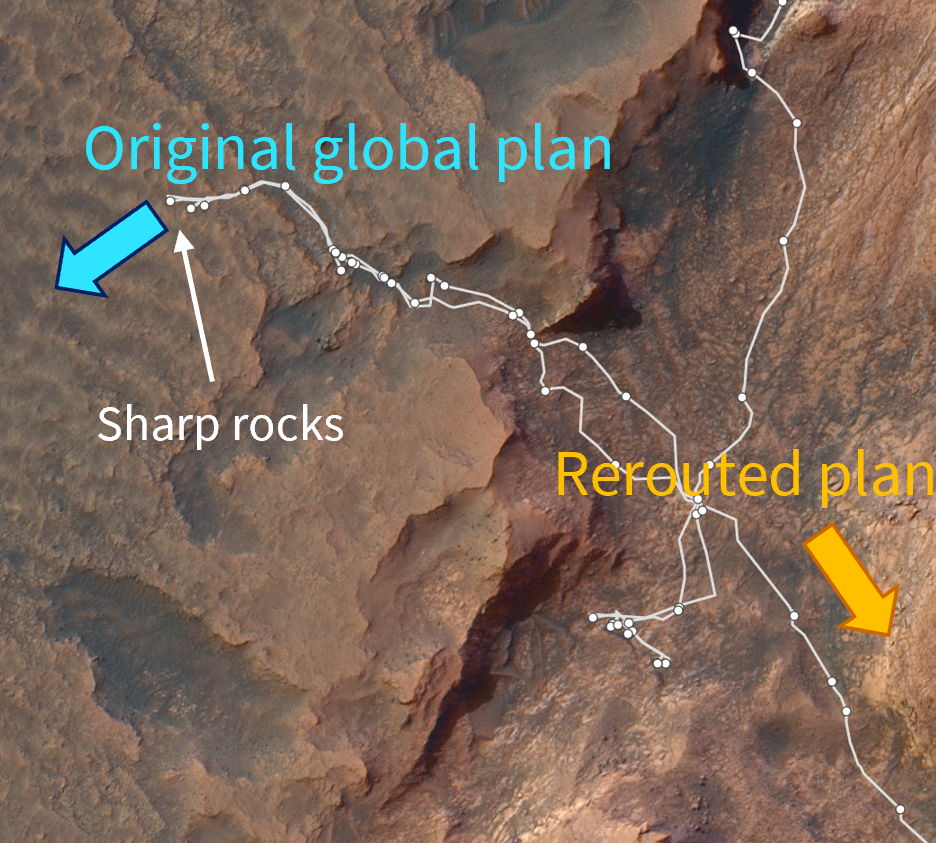}
         \caption{Sol 3345 - 3463.}
     \end{subfigure}
     \hfill
     \begin{subfigure}[b]{0.535\textwidth}
         \centering
         \includegraphics[width=\textwidth]{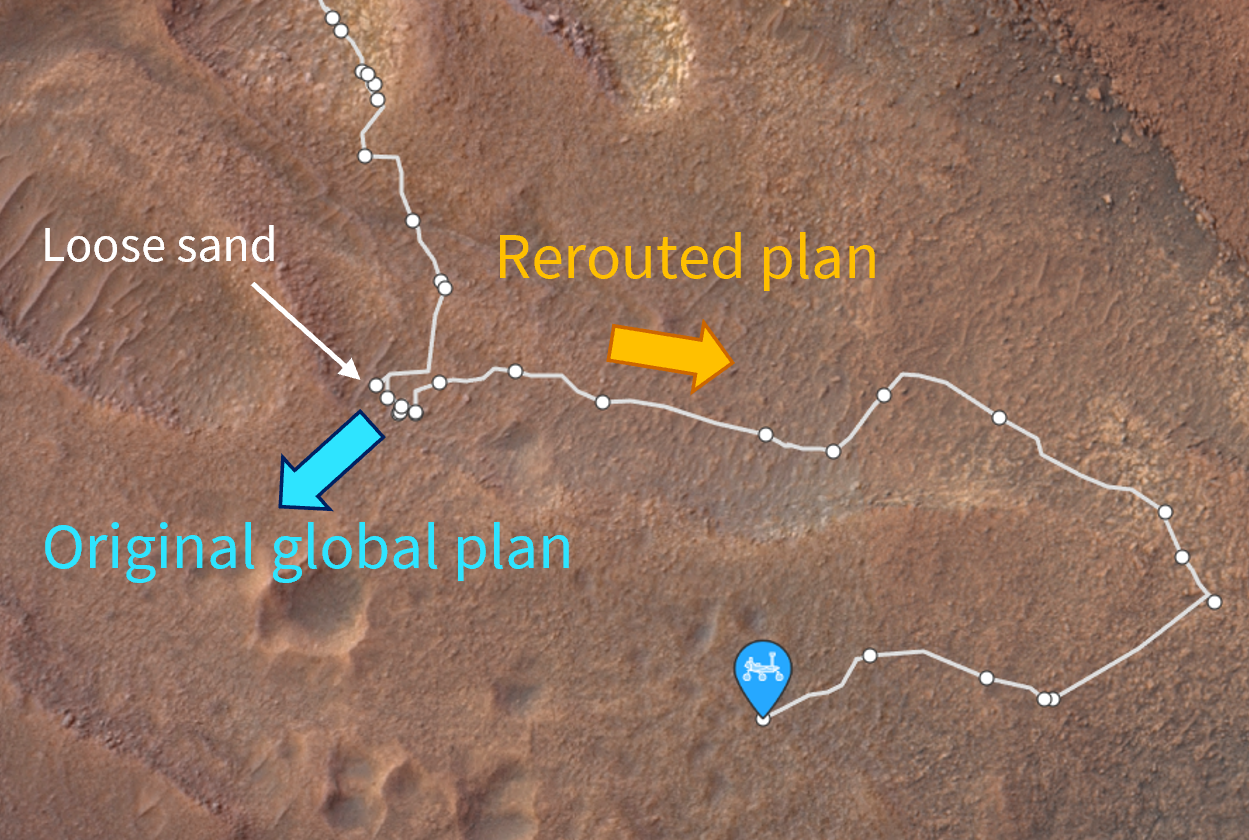}
         \caption{Sol 3815 - 3858.}
     \end{subfigure}
    \caption{Examples of global plan re-routing due to updates from local observations from the Curiosity rover.}
    \label{fig:curiosity_replan}
\end{figure}

However, global paths alone are insufficient for autonomous exploration due to the limited spatial resolution of satellite or drone imagery, coupled with the potential for hazard mischaracterization. 
Figure \ref{fig:curiosity_replan} shows examples from the path taken by the \textit{Curiosity} rover in which the global plan was originally routed through a section of terrain previously thought to be traversable, but then later updated to re-route in a different direction after the rover arrived and the area was deemed untraversable after local observation.
Therefore, it is necessary to develop path-planning algorithms that can integrate both global information and local information procured from rover cameras.

However, onboard navigation algorithms face the challenge of operating within the memory and compute capacity of rover hardware. 
Storing extensive amounts of images and/or topographic data on the rover hardware becomes impractical due to the high memory requirements involved, and association must be performed between images and points of interest which adds computational overhead to onboard processing. 
Neural Radiance Fields (NeRFs)~\citep{mildenhall2021nerf} offer a solution to these challenges by accurately compressing large quantities of visual information and rendering novel views of complex 3D scenes on demand. 
NeRFs can effectively merge environmental information from multiple sources, such as collections of images, topographic data or digital elevation maps. 
Moreover, NeRFs can be incrementally updated to store and reconstruct rover images from previously traversed terrain, further expanding the available global information. 
These advantages make NeRFs suitable for onboard use on planetary rovers.

In this work, we present a novel algorithm for path planning that leverages both local rover images and a NeRF-based representation of the mission site constructed from globally available images. 
These images are the used to construct the NeRF that is passed to the rover. 
Our algorithm combines the local information from rover images and the global information rendered by the NeRF in the vicinity of the rover's location to efficiently construct a traversability cost map and update the planned path. 

The main contributions of our work are:
\begin{itemize}
    \item We introduce a novel NeRF-based method to compress and utilize drone imagery from extraterrestrial environments for path planning. Our method reduces the memory requirements of storing visual information and enables real-time rendering of multiple views of the terrain.
    \item We develop a path planning algorithm that improves efficiency in re-routing based on rover imagery by incorporating global information, such as from the stored NeRF, to improve the reliability of autonomous navigation.
    \item We evaluate our path planning algorithm using high-fidelity simulations in AirSim~\citep{shah2018airsim} with similar terrain to extraterrestrial situations and compare its performance with existing methods that rely either on local or global information sources.
\end{itemize}


The remainder of this paper is organized as follows. 
In Section \ref{sec:prior_work}, we discuss prior work related to rover autonomous navigation, global map information, and Neural Radiance Fields (NeRFs).
In Section \ref{sec:background}, we introduce further background on NeRFs, local path planning, and global path planning.
Section \ref{sec:problem_statement} defines the problem statement, Section \ref{sec:approach} describes the details of our approach, Section \ref{sec:experiments} describes our simulation setup and experiments, and Section \ref{sec:conclusion} concludes this work.
\section{Prior Work}
\label{sec:prior_work}

This section covers relevant prior work on rover autonomous navigation, global information sources for extraterrestrial environments, and Neural Radiance Fields.

\subsection{Rover Autonomous Navigation}
The autonomous navigation system for Mars Exploration Rovers is termed AutoNav~\citep{carsten2007global}, and consists of three main steps.
First, AutoNav uses Grid-based Estimation of Surface Traversability Applied to Local Terrain (GESTALT)~\citep{biesiadecki2006mars, goldberg2002stereo} to process onboard images and generate a local cost map for planning.
The algorithm fits planes to point clouds generated from stereo images to estimate factors such as slope and roughness.
Next, the candidate paths of constant curvature arcs are evaluated based on the cost map, and an arc to execute is selected.
Finally, the rover is driven along the desired arc, and the process is repeated.
More recently, for Mars 2020, and enhanced version of AutoNav termed ENav was developed~\citep{toupet2020ros}.
ENav employs the same general approach as AutoNav, consisting of stereo image acquisition and processing, terrain analysis and path selection, and drive execution, but performs these functions in parallel rather than serially \citep{rieber2022planning}.
These algorithms all focus on local path planning, and do not consider the integration of local and global path planning.

Recent works have also investigated improvements to rover autonomous navigation, using machine learning to aid in path selection.
\citet{abcouwer2021machine} propose a learned heuristic that predicts the outcome of path evaluation checks for a given terrain height map.
This heursitic is then used to influence the cost map to favor paths that would be more likely to pass checks.
In \citet{daftry2022mlnav}, a network is trained to learn intuitively safe paths from human rover driver examples, and the classic search-based planner is augmented with the learned heuristic.
However, both of these works assume standard terrain analysis and cost map generation which relies only on onboard stereo imagery and do not discuss any approaches for integrating global satellite imagery or information into onboard planning.


\subsection{Global Information Sources} \label{sec:global_info}

Generally, global information for extraterrestrial surfaces in available through two channels: surface imagery and laser altimetry. 
Surface imagery can be obtained via orbiters (such as the Lunar Reconaissance Orbiter (LRO) or Mars Reconaissance Orbiter), landers (such as the Apollo landers), aerial vehicles on bodies with atmosphere (such as the Ingenuity helicopter), or from rovers themselves.
Prior work for the Mars 2020 mission~\citep{ono2016data} used Mars Reconaissance Orbiter imagery to determine surface traversability, through a combination of terrain classification, rock density estimation, manual traversability assessment.
Relatively speaking, imagery is easy to capture, but difficult to process and store onboard, with a significant bottleneck being association between local rover images and the surface imagery database.

Laser altimetry is almost exclusively captured from an appropriately instrumented orbiter (Lunar Orbiter Laser Altimeter on LRO or Mars Orbiter Laser Altimeter from \textit{Mars Global Surveyor}).
These altimetry measurements can then be used to generate a digital elevation model (DEM) of the terrain.
\citep{ono2016data} also utilizes a DEM, constructed from stereo processing of images, in traversability analysis.
Although DEMs can provide extensive coverage and globally referenced information, they do not provide visual information, and are expensive to construct and update.

\subsection{Neural Radiance Fields}

Recently, Neural Radiance Fields (NeRFs)~\citep{mildenhall2021nerf} have emerged as a compact yet detailed 3D scene representation.
A NeRF is a neural network which is trained on 2D images, and learns to synthesize novel views of the scene.
The NeRF architecture is designed around the principle of volumetric rendering, which enables it to capture complex visual effects, such as realistic lighting, and fine details.
Since its introduction, have rapidly taken over the computer vision and computer graphics communities, with hundreds of papers and preprints published on the topic since 2020.
In addition to their original application to novel view synthesis, NeRFs also exhibit high compression rates for the input views and enable fast rendering of relevant views at runtime depending on the camera location and orientation. For example, \cite{reiser_kilonerf_2021} show that NeRFs can achieve a compression ratio of up to 5000:1 compared to storing all input views as RGB images, while maintaining high quality in the synthesized views. Similarly, recent work has shown that NeRFs can be rendered at interactive speeds by efficiently leveraging GPU parallelization~\citep{garbin_fastnerf_2021}. These properties enable using NeRFs as compact representations of environment that can be queried on the fly for path planning~\citep{adamkiewicz_vision-only_2022}.  

Although the original NeRF was demonstrated for objects and indoor scenes, recent work has demonstrated its applicability for vast outdoor scenes with satellite scale images \citep{xiangli2022bungeenerf}.
Furthermore, generation of NeRFs for Mars' surface has been demonstrated by MaRF~\citep{giusti2023marf}.
The authors generate NeRFs from images captured by the Curiosity rover, Perseverance rover, and Ingenuity helicopter, and cite NeRFs as a solution for 3D modeling of the Mars surface which circumvents the high compute costs of reconstruction and lack of generalization to unseen scenes from traditional methods.

Recent work has also considered motion planning within NeRFs~\citep{adamkiewicz_vision-only_2022, chen2023catnips}. 
These works directly use the spatial density information of the NeRF to plan trajectories through the environment.
However, NeRF spatial densities are often noisy, with ``clouds" of density in empty space, as the network is trained on an image reconstruction task, with no explicit constraints of the geometric quality of the density itself.
In contrast, we propose using a NeRF a compact representation of visual information to aid in terrain association for costmap updating.









\section{Background}
\label{sec:background}

In this section, we introduce relevant background for Neural Radiance Fields (NeRFs), local path planning, and global path planning. 

\subsection{Neural Radiance Field Training and Rendering}
\label{sec:nerf_background}

\begin{figure}[!ht]
    \centering
    \includegraphics[width=0.8\textwidth]{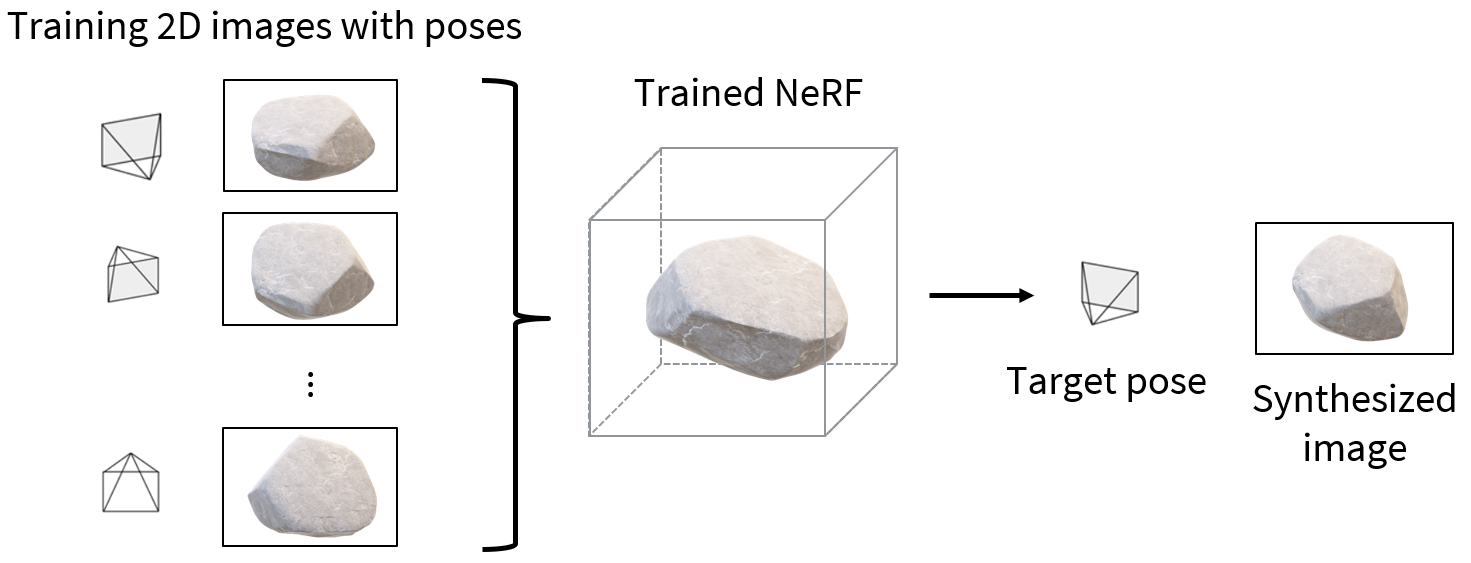}
    \caption{Overview of the Neural Radiance Field (NeRF) training and novel view synthesis task. Given a set of 2D training images with camera poses, the NeRF trains a neural network to represent the scene. Then, given a target camera pose, the NeRF can synthesize a new image from that camera pose of the scene.}
    \label{fig:nerf_setup}
\end{figure}

Here, we will explain more details regarding Neural Radiance Field (NeRF) network architecture, as well as how they are trained and used to render images.
As mentioned in prior work above, NeRFs are trained on a set of posed 2D images, and once trained, can synthesize an image from a novel view specified by arbitrary camera pose.
This is illustrated in Figure~\ref{fig:nerf_setup}.

\begin{figure}[!ht]
    \centering
    \includegraphics[width=0.8\textwidth]{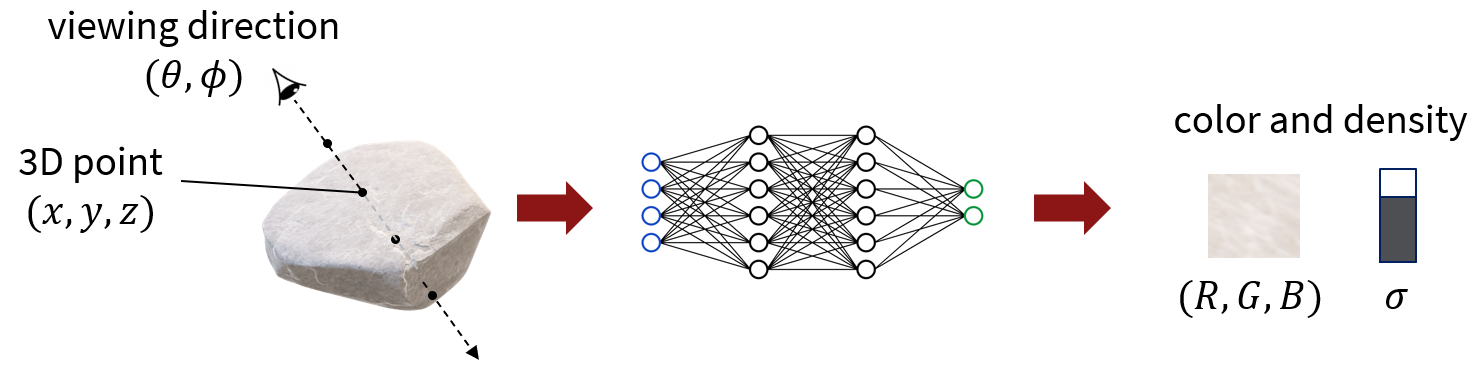}
    \caption{Inputs and outputs of the NeRF neural network. The network takes a 3D point in the scene and viewing direction of the point as input, and learns to predict the color and volumetric density of that point when viewed from that direction. In order to synthesize an image, a NeRF uses multiple rays (viewing directions), and for each ray, queries points along the ray and accumulates their predicted color and densities to compute an RGB pixel value.}
    \label{fig:nerf_network}
\end{figure}

NeRFs accomplish this by learning the rendering function for a given set of data.
In particular, given input 3D coordinate $(x,y,z) \in \R^3$ and viewing angle $(\theta,\phi) \in \R^2$, a NeRF uses a fully-connected neural network to predict output color values $(R,G,B) \in \R^3$ and volume density $\sigma \in \R$ at that point in the scene (illustrated in Figure~\ref{fig:nerf_network}). 

In this paper, we represent a NeRF as a function $F: \R^5 \to \R^4$, where
\begin{equation} \label{eq:nerf}
F(x, y, z, \theta,\phi) = (R,G,B,\sigma).
\end{equation}
This network is trained with a reconstruction loss over the input images.
In order to reconstruct an image from a given target camera pose, we consider a ray for each pixel, with origin at the camera position and associated viewing direction for the pixel.
Then, points are sampled along the ray, and corresponding color and density values are computed from the NeRF network.
These values are then accumulated by weighting the colors by density and summing to obtain the RGB value for the given pixel.

In addition to color, it is also possible to obtain depth information from a given camera pose from the NeRF.
If we consider the density values along a given ray, the values will be low as the ray passes through empty space, and then hit a peak when the ray first passes through occupied space. 
Thus, for a given ray, we can determine the depth by accumulating density, and find the point at which the accumulated density passes a threshold.
We will use these ray accumulation techniques to obtain color and depth in the formation of NeRF terrain features in Section~\ref{sec:nerf_features}.

\subsection{Local Path Planning} \label{sec:local_planning}


We base our local path planning on the practically verified AutoNav algorithm~\citep{carsten2007global}. 
First, an elevation map is extracted from stereo imagery, and each cell is evaluated for traversability to form a local costmap.
Then, an optimal arc is selected from a set of candidate arcs based on traversability cost, steering cost, and distance to goal.

\subsubsection{Local Traversability Cost}
\begin{figure}[!ht]
     \centering
     \begin{subfigure}[b]{0.25\textwidth}
         \centering
         \includegraphics[width=\textwidth]{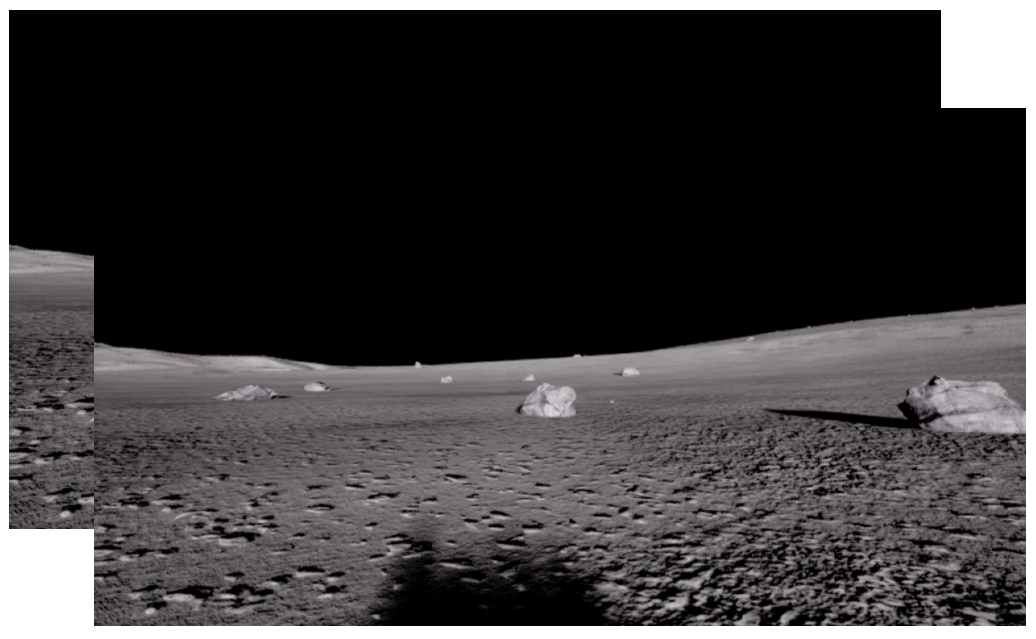}
         \caption{Stereo RGB images from rover.}
         \label{fig:autonav_stereo}
     \end{subfigure}
     \hspace{1.5cm}
     \begin{subfigure}[b]{0.27\textwidth}
         \centering
         \includegraphics[width=\textwidth]{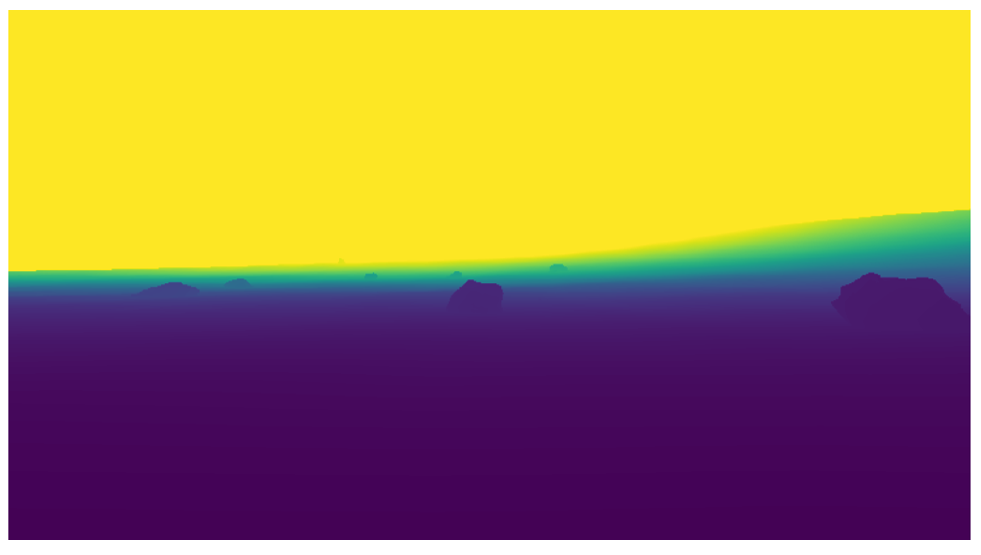}
         \caption{Extracted depth image.}
         \label{fig:autonav_depth}
     \end{subfigure}
     \hspace{0.3cm}
     \begin{subfigure}[b]{0.35\textwidth}
         \centering
         \includegraphics[width=\textwidth]{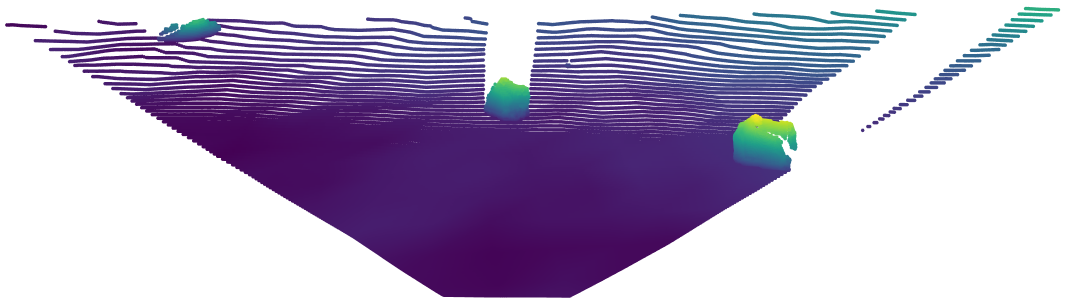}
         \caption{Projected point cloud. }
         \label{fig:autonav_point_cloud}
     \end{subfigure}
     
     \vspace{0.5cm}
     \begin{subfigure}[b]{0.6\textwidth}
         \centering
         \includegraphics[width=\textwidth]{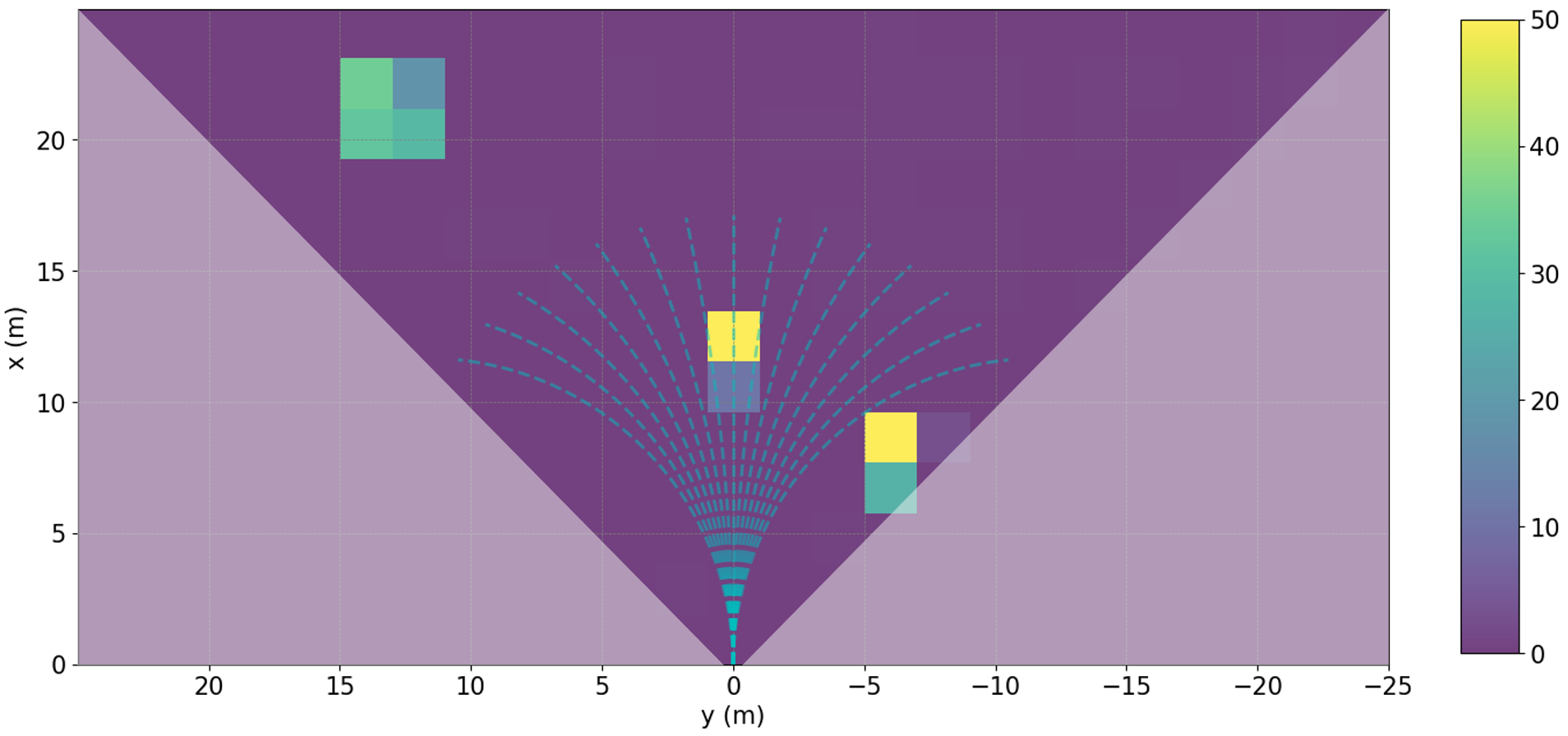}
         \caption{Computed local costmap. }
         \label{fig:autonav_costmap}
     \end{subfigure}
    \caption{AutoNav local costmap pipeline. First, a depth image (\ref{fig:autonav_depth}) is extracted from stereo RGB images from the rover's onboard cameras (\ref{fig:autonav_stereo}). Next, the depth image is projected into 3D to produce a point cloud in local frame (\ref{fig:autonav_point_cloud}). Finally, the 3D space is gridded up into a 2D costmap, where the cost of each cell is determined from the gradient (slope) and variance (roughness) of points within.}
    \label{fig:autonav}
\end{figure}

Stereo RGB images (Figure \ref{fig:autonav_stereo}) are compared to compute a disparity map, and this disparity map is then used to generate a depth image $D \in \R_{+}^{h\times w}$ (Figure \ref{fig:autonav_depth}), where $h$ and $w$ are the height and width of the image in pixels respectively, and each value $d_{ij} \in D$ represents the depth of pixel $ij$.   
Next, the depth map is projected into 3D space using a standard camera projection model in order to obtain a 3D point cloud (Figure \ref{fig:autonav_point_cloud}) in the rover local coordinate frame $P \in \R^{N\times 3}$, where $N = h \times w$.
In practice we also filter out points with a depth greater than threshold $d\thresh$.
The local costmap $C\local \in \R_{+}^{d\thresh \times 2d\thresh}$ (Figure \ref{fig:autonav_costmap}) is then defined as a grid of resolution 1 \si{m} with dimensions $d\thresh \times 2d\thresh$, and its values are computed as the weighted sum of the gradient and variance of the $z$ values of points in $P$ which lie in the grid cell.

\subsubsection{Path Selection}
Once local traversability cost has been computed from local observations, the final step is to select a path for execution.
We use a set of constant speed and curvature arcs as candidate paths, shown overlaid on the costmap in Figure \ref{fig:autonav_costmap}.
These arcs all have constant driving speed $v$, and are parameterized by turning rates $\omega \in [-\omega_\regtext{max}, \omega_\regtext{max}]$.

In order to evaluate the candidate arcs, we use the weighted sum of the three objectives used in AutoNav: steering cost, traversability cost, and global cost.
\begin{enumerate}
    \item Steering cost: simply computed as $\alpha|w|$ (where $\alpha$ is the weighting constant) to incentivize driving straight.
    \item Traversability cost: computed by accumulating costmap values of points sampled along the arc, i.e., $\sum_{i=1}^N C\local(x_i, y_i)$.
    \item Goal cost: computed based as the Euclidean distance from the arc endpoint to the final goal. 
\end{enumerate}
The arc which minimizes this weighted sum is selected as the optimal arc for execution.



\subsection{Global Path Planning} \label{sec:global_planning}

In this work, we refer to global path planning as path planning at a high-level ``global" scale, in which planned paths consist of a sequence of waypoints which may be spaced out by multiple rover lengths, and do not necessarily consider rover dynamics.

This flavor of path planning has been studied extensively, with the standard approach involving minimum cost path search over a graph.
The graph represents state space, and often constructed by simply gridding 2D space into cells, where cells are nodes, and adjacent cells are neighbors.

\subsubsection{Global Traversability Cost}

The primary sources of data to inform terrain traversability at the global scale are satellite imagery and laser altimetry (see Section \ref{sec:global_info}).
Detailed satellite imagery, such as from MRO HiRISE (approx. 1 \si{m} per pixel resolution) or LRO (0.5 \si{m} per pixel), can identify large scale features such as craters and sloped terrain, as well as approximate terrain characteristics. 
However, features smaller than the image pixel resolution, such as small rocks or loose sand, cannot be resolved.
For the Mars 2020 mission, HiRISE imagery was analyzed for a combination of terrain classification, slope, rock density, regions of interest, and known hazards to perform traversability analysis and inform the eventual landing site selection~\citep{ono2016data}.
On the other hand, laser altimetry is crucial for creating accurate digital elevation models (DEMs), which can be used to inform slope analysis for traversability.

As our work focuses more on how to update the global costmap based on local observations, the details of global costmap initialization are outside the scope of our work, and we employ a basic strategy based on pixel intensity values of a top-down image to initialize our global costmap.
For a realistic implementation, the global costmap would be initialized using techniques such as from \citet{ono2016data}, but due to limitations in global information sources we would still expect there may be regions in which traversability is overestimated (such as sharp rocks or loose sand, see Figure~\ref{fig:curiosity_replan}).

\subsubsection{Minimum Cost Path Search}

Once a global costmap has been constructed, a graph search algorithm such as A* or Field D*~\citep{ferguson2007field} is used to compute the global path to a desired goal as a set of waypoints.
For our implementation, we use A*, with inputs to the planner being the current rover location and desired goal location.
We use four-directional node neighbors, and Euclidean norm distance as the heuristic cost estimate.
\section{Problem Statement}
\label{sec:problem_statement}

We consider the problem of navigating a planetary rover to a desired goal location while avoiding obstacles and untraversable regions, and minimizing an unknown scalar cost $C^*$ representing critical factors such as traversability, time, energy, and wear and tear on the rover's systems. 
Additionally, the rover is equipped with an onboard stereo camera, which is used to capture imagery and determine the local cost $C\local$ in the vicinity of the rover, as described in Section~\ref{sec:local_planning}.

In our mission environment, we operate under the assumption that global information is readily accessible through aerial imagery obtained via drone reconnaissance. 
This assumption is consistent with established extraterrestrial mission protocols, as evidenced by mission to Mars, where the deployment of reconnaissance drones for surveying potential mission sites before rover deployment is a topic of research interest in the field~\citep{tzanetos2022ingenuity}. 
For the remainder of this paper, the mission environment over planned autonomous drive is referred to as the \textit{global} region.

Using the global information, we initialize the global costmap $C\globl$ and determine the associated global path $P$. 
This can be done using techniques such as described in Section~\ref{sec:global_planning}. 
While the global costmap assigns a cost value $\mathcal{C}\globl(x, y)$ to each surface point $(x, y)$ within the global region, the accuracy of this cost can be compromised by various sources of error in the global information, including factors such as low resolution, adverse viewing conditions, or reliance on outdated imagery. 
Hence, we update the global costmap $C\globl$ with the local cost $C\local$ derived from rover images. This local cost offers the advantage of being closer to the true cost $C^*$ since it is based on the rover's perspective. 
However, it is exclusively available for the small portion of the global region that falls within the rover's current field of view, referred to as the \textit{local} region. 

In this paper, our primary aim is to harness the complementary strengths of the local costmap $C\local$ and the global costmap $C\globl$ to optimize the rover's path dynamically. We achieve this by continuously enhancing the global costmap $C\globl$ with the precision of local cost values $C\local$, ensuring a globally accessible and accurate representation of costs for all points within the global region.

To extend the updating of the global costmap beyond the local region, we employ a NeRF-based representation of the mission site coupled with a feature-based interpolation algorithm. Combined together, this enables each local cost values to influence extensive portions of the global costmap, thereby facilitating efficient re-planning in areas determined to have high traversability costs.

We adopt the path planning hierarchy of local and global planning, with both levels following the paradigm of costmap-based planning \textemdash\ at both levels, the path or trajectory is selected by minimizing the cost over a discretized cost grid. Details on the methods used for local and global planning are provided in Sections~\ref{sec:local_planning} and \ref{sec:global_planning}.






\section{Approach}
\label{sec:approach}

\begin{figure}[!ht]
    \centering
    \includegraphics[width=0.8\textwidth]{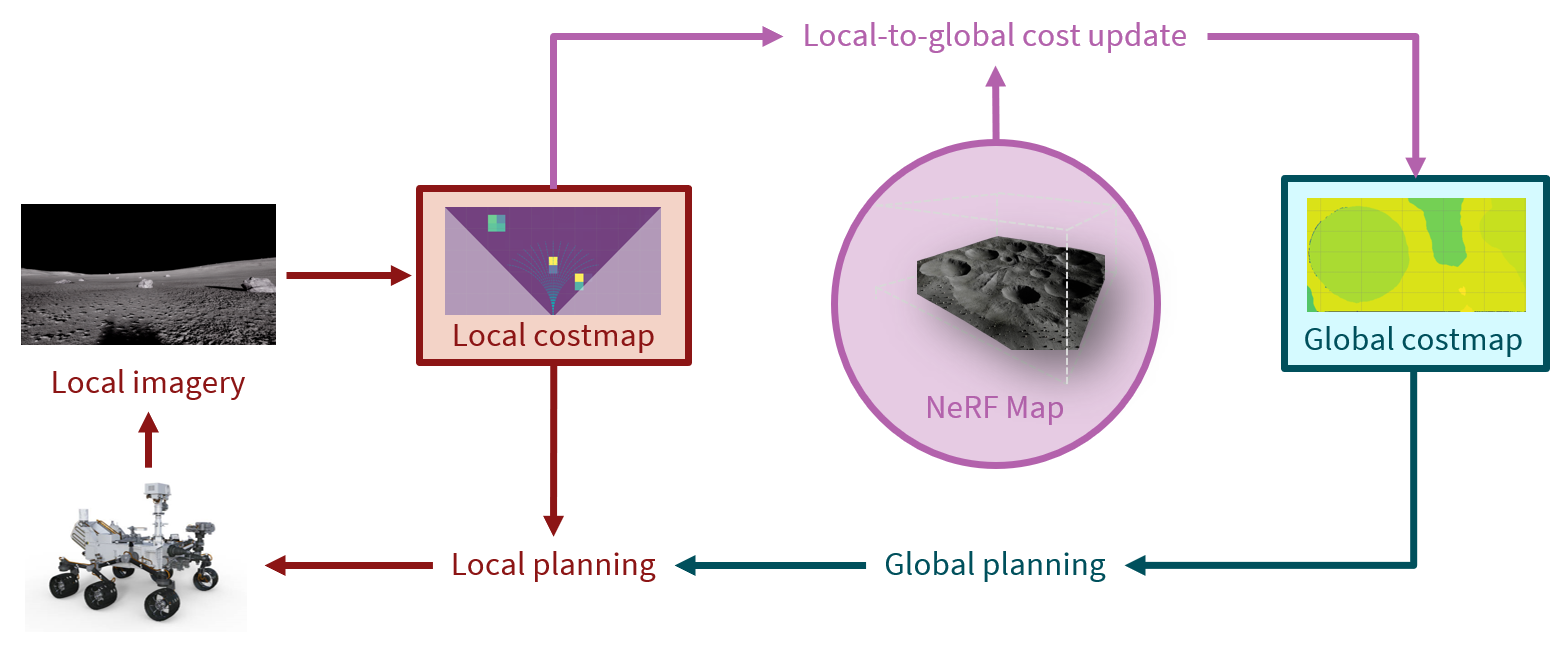}
    \caption{Overview of online planning framework. The rover uses a global costmap to plan a global path for navigation. The global costmap is continuously updated through onboard imagery processing, resulting in the creation of a local costmap which reflects the current terrain and obstacles around the rover. The rover effectively integrates the local costmap into the global costmap using a NeRF-based global representation, facilitating the adaptation of the local cost values to multiple regions within the global costmap. The rover then combines the new global path with the local costmap values to determine a local path for execution.}
    \label{fig:overview}
\end{figure}

Our approach consists of two primary steps: offline NeRF global map generation, and online rover path planning. 
First, a NeRF is trained from prior global imagery, which serves as a map representation of terrain.
The NeRF can be stored in compact form and the weights of the network are uploaded the rover prior to the mission.
Then, during rover autonomous navigation, the rover runs global and local path planning to navigate towards the goal while minimizing path cost.
Figure~\ref{fig:overview} shows an overview of the online planning framework.
 
\subsection{NeRF-based Global Representation of Terrain}
\label{sec:nerf_map}

To enhance the rover's understanding of the mission environment for effective online planning, we employ a NeRF-based representation of the environment. The NeRF is trained using simulated drone images of the mission site, and compactly captures visual and structural details of the terrain that can be queried for specific points on-demand and in real-time. Using this representation, we then develop a feature extraction methodology to compute a compact feature vector representing the structural and visual details associated with any point on the surface. Leveraging these features, we develop a method for effectively updating the global costmap through feature-based interpolation of the local cost values obtained via rover's onboard images (detailed below in Section \ref{sec:local_to_global}).

\subsubsection{NeRF Training}
\label{sec:nerf_training}

To produce a NeRF for a terrain, we primarily rely on images taken along camera trajectories simulated in Unreal Engine via AirSim~\citep{shah2018airsim}. 
The simulation mimics a Lunar-like surface characterized by features such as craters, rocks, and varying elevations. These trajectories trace an approximate pattern of concentric circles, at altitudes ranging from 10~\si{\meter} to 200~\si{\meter}.

Unreal Engine is our chosen simulation tool because it facilitates a realistic representation of extraterrestrial terrains, both visually and in modeling the physical behavior. This capability is essential for our path planning application, as well as AirSim's ability to simulate vehicle movement within the environment.

For each trajectory, we gather $200$ images at a resolution of $720 \times 480$ pixels. The camera is set at a $45$-degree viewing angle, facing the terrain during the drone's flight. This specific trajectory design ensures that the terrain is captured from various angles, enabling a detailed understanding of the 3D structure underlying the surface.

In practice, accurate ground truth pose data isn't accessible when operating drones on extraterrestrial surfaces. Therefore, we employ COLMAP~\citep{schonberger2016structure} to establish the pose. COLMAP achieves this by densely matching SIFT features~\citep{suzuki20113d} across the collected images in a post-processing step. 

For NeRF training, we utilize the Nerfstudio~\citep{tancik2023nerfstudio} framework, with a specific focus on the Nerfacto model. This model is a combination of several published methods proven effective for real data, including the ability to work well with unbounded scenes, static scene captures, and a variety of camera perspectives. Future work will explore developing NeRF models that utilize the unique characteristics of extraterrestrial terrains to improve performance.

\begin{figure}[!ht]
    \centering
    \includegraphics[width=0.75\textwidth]{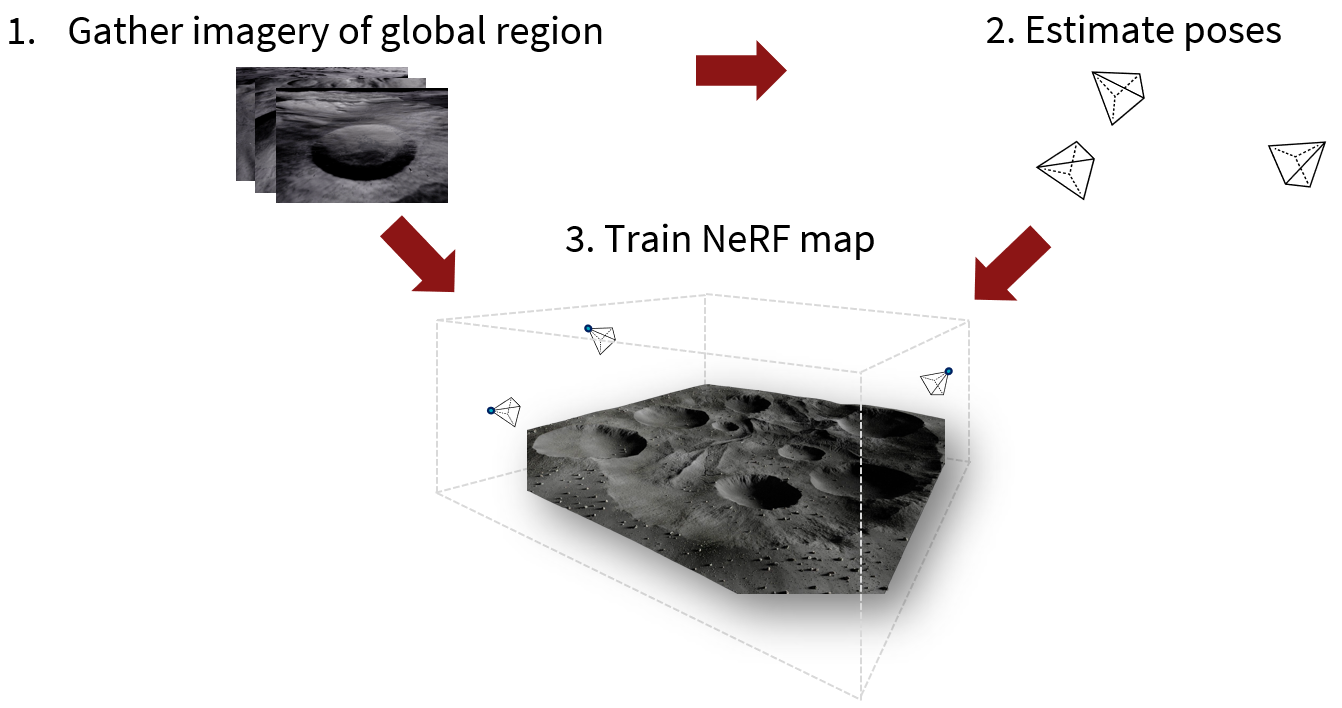}
    \caption{Neural Radiance Map construction process. First, imagery of the global region is gathered from simulated camera trajectories. Next, poses of these images are estimated, and the images and poses together are used to train a NeRF of the global region.}
    \label{fig:nerf_training}
\end{figure}

\subsubsection{NeRF Features}
\label{sec:nerf_features}

Once we have trained our Neural Radiance Map of the region of interest, we would like to use it as a source of global information at runtime for onboard rover re-planning.
We accomplish this by introducing a feature function $\Phi(\cdot)$, which, for a given point $\x$ in 2D space, allows us to extract a feature vector $\Phi(\x)$ which serves as a descriptor for the local terrain at $\x$.
These features then allow us to compute similarity across patches of terrain in order to interpolate costs (Section \ref{sec:local_to_global}).

The feature function $\Phi$ is computed as follows.
For patch of 2D space centered around coordinate $\x$, we render $N\rays$ downward-facing rays.
For each ray $i$, we sample RGB and density values, which are accumulated to obtain color and depth $(\rgb_i, d_i) \in \R^4$ (refer to Section~\ref{sec:nerf_background} for more details on the ray accumulation process).
These values are then concatenated to form the output feature vector:

\begin{equation}
    \Phi(\x) = 
    \begin{bmatrix} \rgb_1 \\ d_1 \\ \vdots \\ \rgb_{N\rays} \\ d_{N\rays} \end{bmatrix},
    \Phi(\x) \in \R^{4N\rays}.
\end{equation}

\begin{figure}[!ht]
    \centering
    \includegraphics[width=0.75\textwidth]{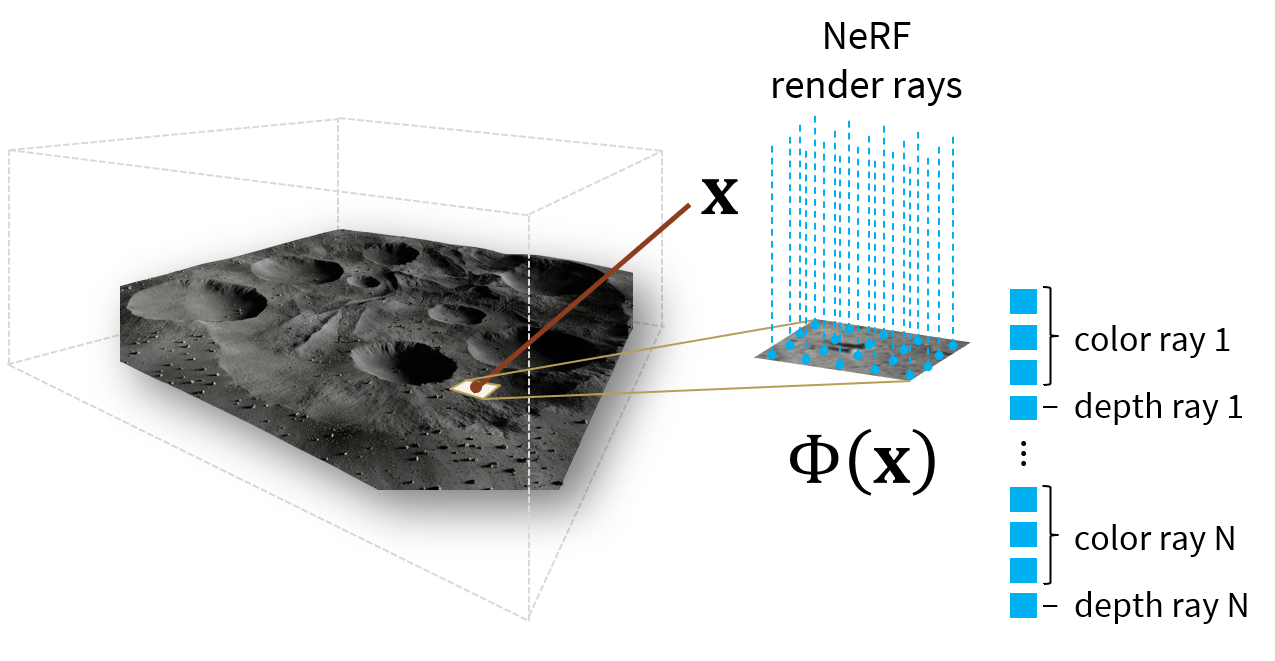}
    \caption{NeRF terrain feature extraction. For a given point in the scene, we render a group of rays in a surrounding patch of terrain. For each ray, we accumulate color and densities from the NeRF to obtain both color and depth values. We then concatenate all the values together into a single vector, which forms the feature vector for given point.}
    \label{fig:nerf_features}
\end{figure}

This feature extraction process is illustrated in Figure~\ref{fig:nerf_features}.


\subsection{Online Local-to-Global Re-planning}
\label{sec:local_to_global}

Now, we return to our overall planning framework (Figure \ref{fig:overview}). 
After we have taken local sensor observations to form a local costmap, we also want to propagate information from them to update our global costmap. 
A naive way to do this would be to simply fill in local cost values into the global costmap. 
However, local cost values are sparse, we wish to make maximum use of the information and efficiently update as much of the global map as we can.
For example, if we observe the edge of a cluster of rocks, we would like to mark the entire area as high cost in the global map and ``globally" re-route, rather than proceeding forward and continually ``locally" re-routing.

Our key idea is, for each global cluster, use kernel ridge regression (KRR) to interpolate the local cost values over the larger global region.
In this way, we propagate the information contained in sparse local cost observations over the larger global region to enable global re-planning around newly observed terrain.

\subsubsection{Local Cost Samples}

As a first step, we need to convert local cost observations from the local costmap into the global coordinate frame.
The local costmap $C\local$ is converted to a collection of point samples ${(\x_i^{\regtext{local}}, c_i)}$ in local frame, where $\x_i^{\regtext{local}} \in \R^2$ are sampled from the costmap grid, and $c_i = C\local(\x_i^{\regtext{local}}) \in \R$.
Then, the sample positions $\x_i$ are converted from local to global coordinates using knowledge of the rover pose, and we extract features $\Phi(\x_i)$ using the NeRF, which we will use for interpolation in the following step.


\subsubsection{Cost Interpolation} \label{sec:cost_interpolation}

We use kernel ridge regression (KRR)~\citep{vovk2013kernel} with a linear kernel to interpolate global cost values from ``local" samples $(\Phi(\x_i), c_i)$ obtained above.
KRR is a commonly used regression technique over high-dimensional feature spaces, which matches well with our choice of the NeRF feature space as the interpolation domain.


First we introduce a global cost function $\mathcal{C}\globl : \R^2 \rightarrow \R_{+}$, such that $\mathcal{C}\globl(\x)$ is the estimated global cost of location $\x$.
We parameterize $\mathcal{C}\globl(\x)$ as a linear function $\w^T \Phi(x)$, where $\w \in \R^{4N\rays}$ is a weight vector computed by minimizing the objective
\begin{equation}
    \frac{1}{2} \sum_i (c_i - \w^T\Phi(\x_i))^2 + \frac{1}{2} \lambda \norm{\w}^2,
\end{equation}
which is a least squares objective fitting $\mathcal{C}\globl(\x_i)$ to the $c_i$ with regularization term scaled by $\lambda \in \R$, which prevents overfitting in the high-dimensional feature space.

\subsubsection{Global Region Clustering}

Because the validity of our KRR cost interpolation decreases with spatial distance, we restrict our interpolation to within global clusters, which are determined at initialization.

To initialize the global costmap $C\globl$, we render top-down image $I\globl \in \R_{+}^{H\times W}$ which covers the entire global region, and perform semantic segmentation into pixel-wise clusters using the Segment Anything Model~\citep{kirillov2023segment}. 
These clusters correspond to regions of visual similarity, within which local costs may be interpolated to update global costs.

\section{Experiments and Validation}
\label{sec:experiments}

This section details the experimental validation for our approach. 
We first analyze the reconstruction quality and storage and computation costs of NeRF-based maps trained on terrain imagery.
Next, we implement our full online planning method in real-time, high-fidelity simulation, and demonstrate that our approach generates more desirable paths compared to an approach which does not integrate local and global planning\footnote{Our codebase is available online at \url{https://github.com/adamdai/rover_nerf_planning}.}.

\subsection{NeRF Terrain Reconstruction}

First, we validate the utility of NeRFs for constructing high-fidelity 3D global representations of extraterrestrial landscapes, as outlined in Section \ref{sec:nerf_training}. In addition to the Unreal Engine Lunar-like surface described in Section \ref{sec:nerf_training}, we evaluate NeRF reconstructions of the Martian surface using Google Earth Studio (GES). 

For both environments, we simulate aerial images of the terrain, captured from varying altitudes. In the GES environment, we directly acquire the camera poses through the interface instead of estimating them from the images. 

\subsubsection{Memory Footprint}

Our trained Unreal Engine Moon NeRF model, generated over an approximately 100 m $\times$ 300 m area of extraterrestrial landscape, occupies a memory size of 248 MB.
Notably, the memory footprint of the NeRF model remains constant despite varying the landscapes and their size, affirming that the memory consumption is agnostic to the spatial dimensions and visual details present within the scene.

\subsubsection{Qualitative Evaluation}

\begin{figure}[!ht]
    \centering
    \includegraphics[width=0.9\textwidth]{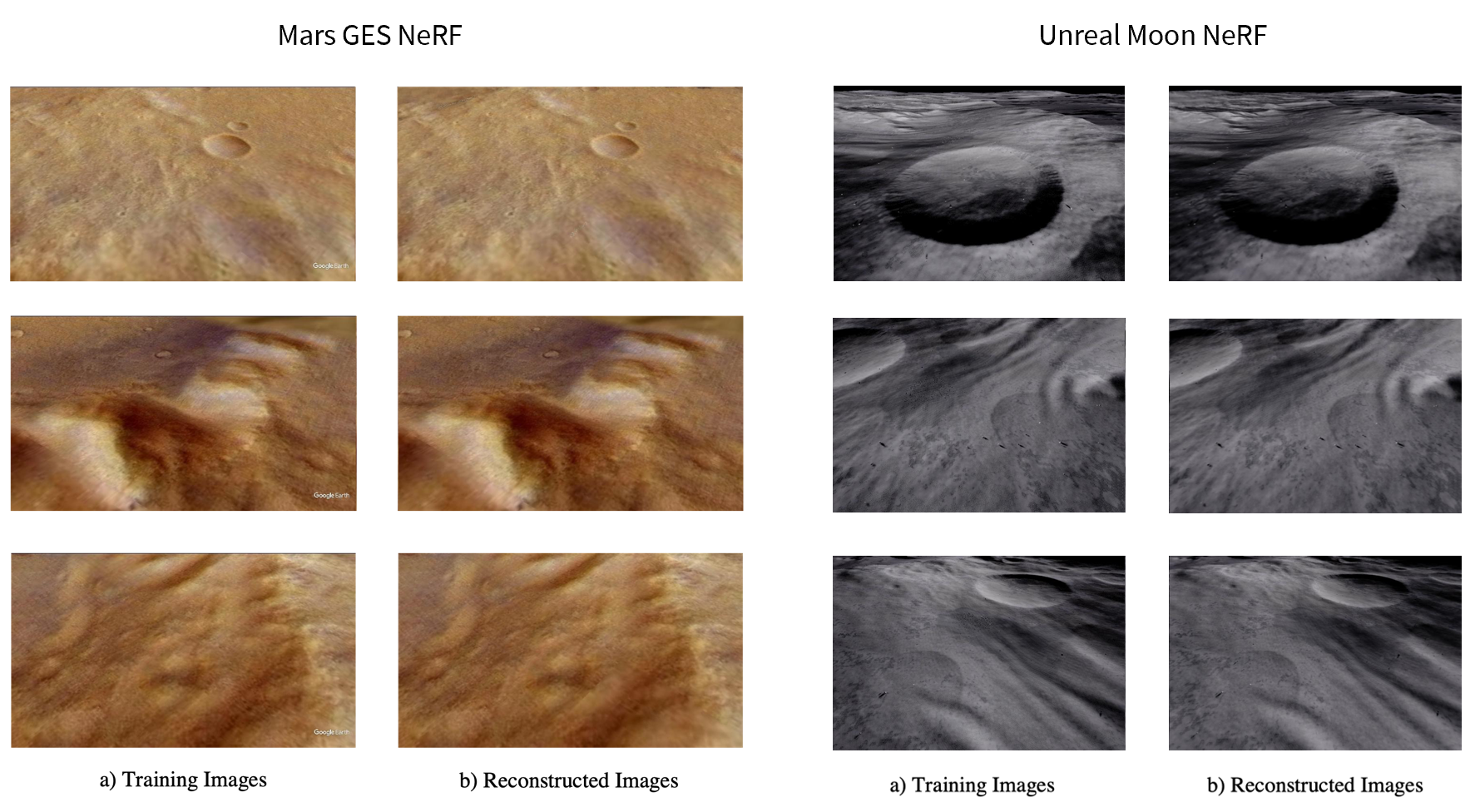}
    \caption{Comparative visualization of NeRF renders and original imagery for both simulated Mars GES data and Unreal Engine Moon environment. The original training images are visualized alongside their NeRF renders, which possess a high degree of similarity to the original images.}
    \label{fig:mars_nerf}
\end{figure}

Figure~\ref{fig:mars_nerf} illustrates example NeRF renders alongside original training images. Visually, our NeRF renders display a high degree of similarity to the original training images.

\subsubsection{Quantitative Evaluation}

To further quantify the quality of our 3D reconstructions, we employ several evaluation metrics: 

\begin{enumerate}
    \item Peak Signal-to-Noise Ratio (PSNR): Our Mars model achieved a PSNR value of 30.77 \si{dB} and our Unreal Engine Moon model achieved a value of 27.83 \si{dB}. This falls close to the typical reference value of 30 \si{dB} for image compression quality between. This suggests a satisfactory level of color reconstruction.
    \item Structural Similarity Index Measure (SSIM): The Mars model measured an SSIM score of 0.954, and Unreal Engine Moon model measured 0.801, which are both close to the maximum value of 1, indicating a high level of structural similarity between the NeRF renders and the original images. This score reaffirms that the geometric features are well-preserved in our 3D representation.
    \item Learned Perceptual Image Patch Similarity (LPIPS): The LPIPS metric was observed to be 0.114 and 0.233 for the Mars and Moon models respectively. Here, a lower score is better, implying a perceptually close resemblance between the NeRF renders and the original images.
\end{enumerate}

Collectively, these metrics validate that our NeRF-based 3D representations of extraterrestrial landscapes are both visually and numerically faithful to the original imagery. This demonstrates the efficacy of our approach in constructing high-fidelity 3D global representations using a variety of data sources.

\subsection{Online Planning Simulation}

Due to the impracticality of real-world testing of navigation on extraterrestrial terrain, we use simulation to test our planning pipeline.
Simulation also allows us to efficiently stress test our approach over many trials and varying terrain and settings.
For future work, we plan to conduct real-world tests with a miniature all-terrain rover at locations with varied and challenging terrain features which mimic conditions on extraterrestrial surfaces.

\subsubsection{Simulation Environment}

\begin{figure}[!ht]
     \centering
     \begin{subfigure}[b]{0.53\textwidth}
         \centering
         \includegraphics[width=\textwidth]{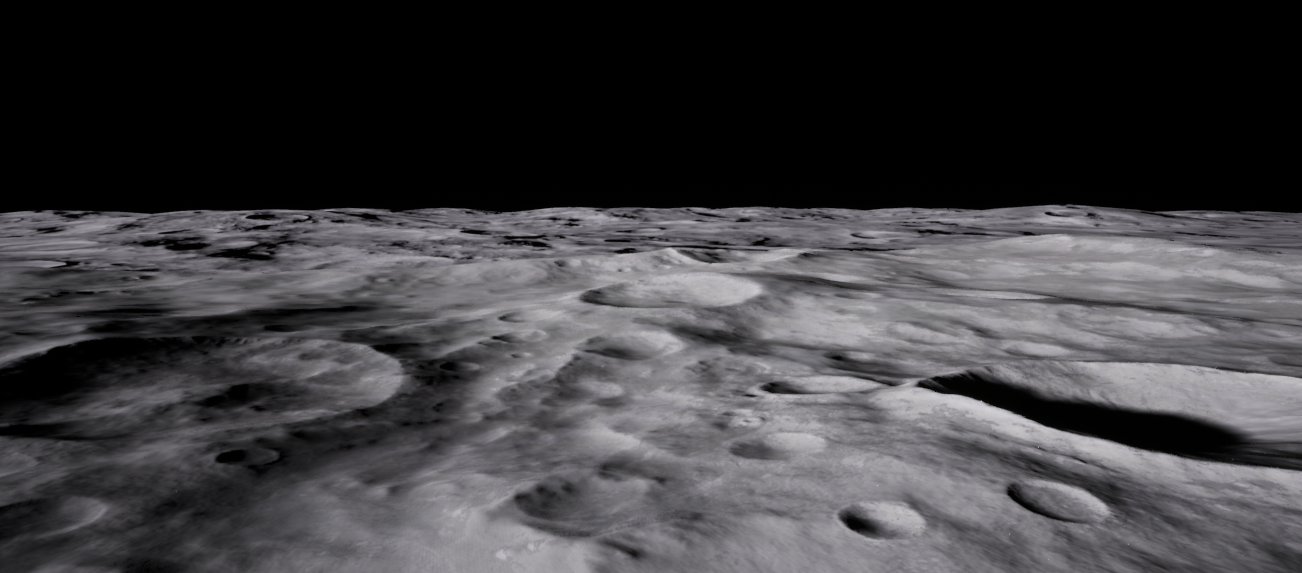}
         \caption{Cinematic view of landscape (captured from within Unreal Engine).}
         \label{fig:moon_landscape}
     \end{subfigure}
     \hfill
     \begin{subfigure}[b]{0.42\textwidth}
         \centering
         \includegraphics[width=\textwidth]{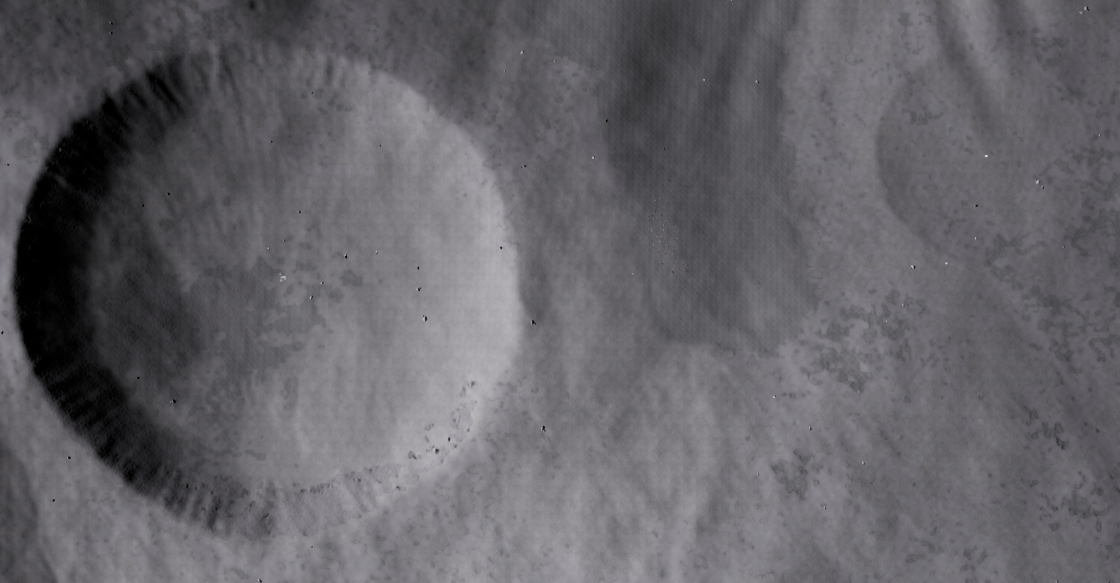}
         \caption{Top-down view of scenario.}
         \label{fig:test_scenario}
     \end{subfigure}
    \caption{Moon simulation environment. The environment features realistic lighting and contains highly detailed $4 \times 4$ km with craters and rocks.}
    \label{fig:moon_env}
\end{figure}

For our simulation environment, we use a ``Moon Landscape" Unreal Engine environment~\citep{karadayi}, shown in Figure \ref{fig:moon_landscape}.
The environment features $4 \times 4$ km stretch of landscape with detailed craters, rocks, and textures, and realistic lighting from a simulated Sun light source.
We use AirSim~\citep{shah2018airsim} to simulate a rover vehicle in real-time within the Moon environment.
AirSim allows us to control the rover with throttle and steering rate commands and render onboard camera imagery.
The simulation handles realistic vehicle physics involved with driving on sloped or rough terrain.

Our algorithm runs global and local re-planning every 5 seconds.
First, stereo imagery is acquired, and a local costmap computed as in Section \ref{sec:local_planning}.
Then, the global costmap is updated as in \ref{sec:local_to_global}, and the global path is re-planned as in \ref{sec:global_planning}.
Finally, we local path is planned according to the local costmap and updated global path, and steering and throttle commands are sent to drive along the selected arc.
Arcs are parameterized by speed of $v = 3.5$ \si{m/s} and turning rates of $\omega \in [-0.3, 0.3]$ \si{rad/s}.

\subsubsection{Planning Simulation Experiments}

\begin{figure}[!ht]
    \centering
    \includegraphics[width=0.95\textwidth]{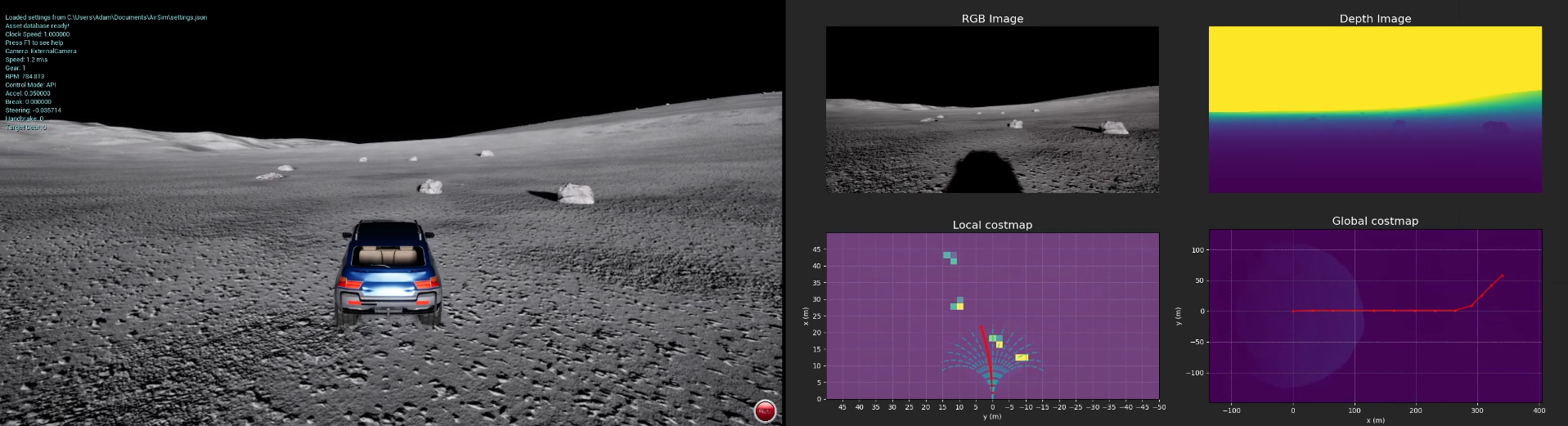}
    \caption{Screenshot of our simulation running in real-time. On the left, the vehicle is shown in the environment, and on the right, the latest RGB image and depth image are shown on top, and the local and global costmap with planned paths overlaid are shown on the bottom.}
    \label{fig:rover_sim_sc}
\end{figure}

For our experiments, we initialize the rover at starting location $(x\start, y\start)$ and specify a high-level goal location $(x\goal, y\goal)$.
A bounding box region with buffer of 100 \si{m} is then defined as the global region, imagery is captured from a spiral camera trajectory as described in Section \ref{sec:nerf_training}, and a NeRF map of the region is trained.
Then, we run our planning pipeline to autonomous navigate the rover through the environment until it reaches within $10$ \si{m} of the goal location.
Figure \ref{fig:rover_sim_sc} shows a screenshot of our simulation running in real-time.

We consider two scenarios for testing: ``medium rocks" and ``small rocks." 
Both scenarios use the global region shown in Figure~\ref{fig:test_scenario}, and share the same start and goal location with a high-cost region in between.
In the first scenario, the high-cost region contains medium-sized rocks ranging between 0.2 to 0.4 \si{m} in approximate diameter, and in the second scenario, it contains small rocks ranging between 0.1 to 0.2 \si{m}.
The small rocks are traversable by the simulated vehicle, but would potentially cause wear and tear in a realistic setting.
On the other hand, the medium rocks are too large to be traversed by our simulation vehicle.

For comparison, we run our full re-planning approach against a baseline approach which performs no global re-planning.
This baseline corresponds to the strategy of running AutoNav or standard local planner while following a fixed global plan for extended autonomous drive.
For evaluation, we use total path cost, total path length, total elapsed time, and number of collisions with rocks as metrics.
Path cost is computed by accumulating the local cost from the traversability costmap computed during local path planning. 


\subsubsection{Planning Simulation Results}

\begin{figure}[!ht]
     \centering
     \begin{subfigure}[b]{0.95\textwidth}
         \centering
         \includegraphics[width=\textwidth]{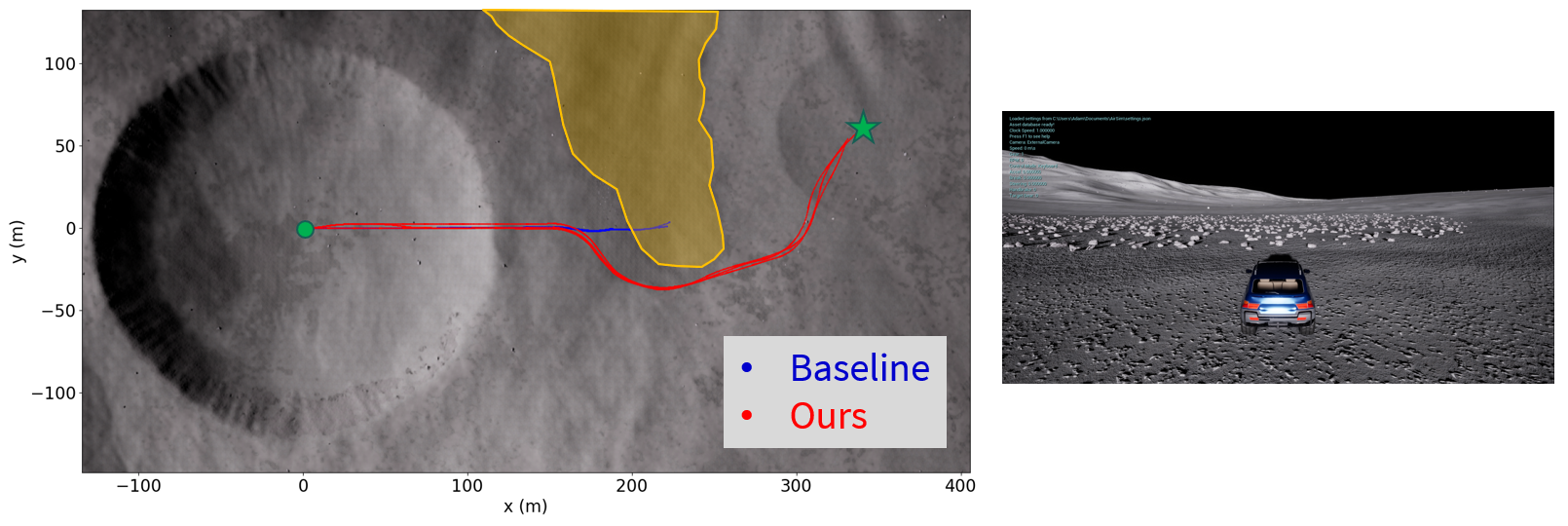}
         \caption{Medium rocks scenario. The baseline approach does not update the global plan after the untraversable rocks region is observed, and continues to route the rover through, causing it to eventually become stuck in all 3 of 3 of the runs. On the other hand, our approach uses local observations to update the global costmap, and thus when the untraversable region is observed the global path is updated accordingly to route around it and eventually reach the goal.}
         \label{fig:med_rocks_paths}
     \end{subfigure}
     \begin{subfigure}[b]{0.95\textwidth}
         \centering
         \includegraphics[width=\textwidth]{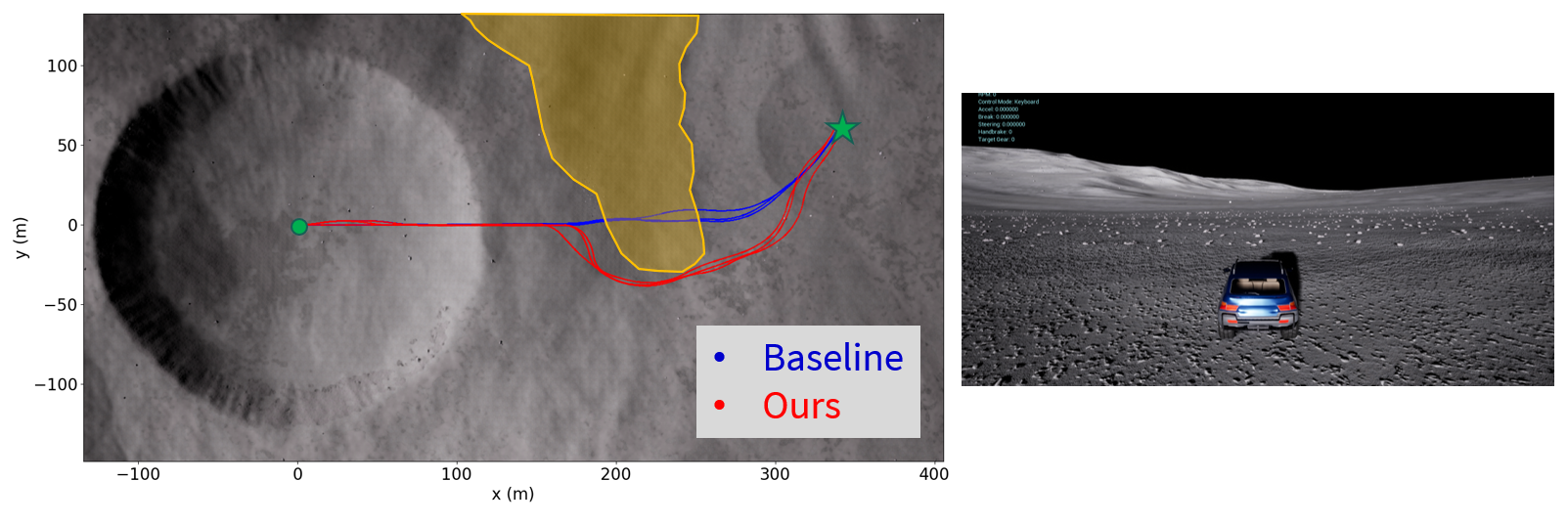}
         \caption{Small rocks scenario. Similarly, we observe the baseline approach routes the rover through the rocky region while our approach replans around it. In this scenario, the rover is capable of traversing the rocks, but it accumulates cost in doing so, which is reflected in the path metrics shown below in Table 1.}
         \label{fig:small_rocks_paths}
     \end{subfigure}
    \caption{Comparison of paths taken by baseline approach vs. our approach for two scenarios. Both scenarios share the same global region and start and goal locations, marked by a green circle and start respectively, and contain a high-cost region, highlighted in yellow. In the first scenario (top), this high-cost region is filled with medium sized rocks, and the second scenario (bottom) features small rocks. In both scenarios, our approach (paths shown in blue) successfully routes the rover around the high-cost region after it is observed, whereas the baseline approach (paths shown in red) paths the rover through the high-cost region, causing it to become stuck and/or accumulate high-cost. }
    \label{fig:path_results}
\end{figure}

Here, we present our results for the two scenarios comparing our approach to the baseline approach.
Figure~\ref{fig:path_results} shows a comparison of the paths taken by rover using the baseline approach versus paths using our approach. 
In both scenarios, we observe the baseline is unable to route the rover around the high-cost rock field, as even though the rover locally observes the rock field as high cost upon reaching the region, and attempts to locally replan accordingly, the global plan stays fixed, and continues to route the rover through the rock field.
Additionally, while the local planner is able to handle routing the rover around sparse rocks, the rock field is too large of an obstacle for the local planner to adequately handle.
In contrast, under our planning algorithm, when the rover observes the rock field with its onboard cameras and classifies it as high-cost, those cost values are then placed in global frame and a global cost function is regressed over them, which results in the entire cluster receiving higher cost in the global costmap, and the global path being re-routed around the rock field.

Table 1 shows a comparison of path metrics of the baseline vs. our approach for the small rocks scenario.
We ran our simulation for each approach three times, and averaged total path cost, number of collisions, total time, and total path distance.
The metrics show that, although the baseline approach resulted in shorter planned paths in terms of time and distance, the paths resulted in collisions with the small rocks, and accumulated higher cost, whereas our approach planned paths with no collisions and lower cost, as a result of steering clear of the high-cost rock field region.

\begin{table*}[htbp]
\label{tab:path_metrics}
    \caption{Comparison of path metrics for baseline and our approach (averaged over three runs each). The baseline approach plans paths through the global high-cost region, resulting in shorter paths with higher cost and collisions, whereas our approach re-routes the global plan on the fly, resulting in slightly longer paths but with lower cost and no collisions.}
    \begin{center}
    \begin{tabular}{|c|c| c|}
        \hline
        & \textbf{Baseline} & \textbf{Ours (replan)} \\
        \hline \hline
        Total path cost      & 39.77        & 13.44        \\
        Number of collisions & 4.3          & 0            \\
        Total time           & 289.6 \si{s} & 323.6 \si{s} \\
        Total path distance    & 335.7 \si{m} & 378.6 \si{m} \\
        \hline
    \end{tabular}
    \end{center}
\end{table*}

\pagebreak
\section{Conclusion}
\label{sec:conclusion}

In this work, we consider the problem of autonomous rover navigation, and utilize Neural Radiance Fields (NeRFs) as a global map representation.
We develop a planning framework which integrates local and global path planning, using the NeRF map to propagate local cost information to the global costmap.
This allows the algorithm to re-route the rover around previously unseen high-cost regions based on onboard observations, resulting in improved adaptivity in path planning over longer autonomous drives.

Going forward, we hope this work will pave the way for future applications of NeRFs to extraterrestrial navigation and path planning.
In particular, we aim to develop custom NeRF architectures for terrain modeling, as well as explore alternative global cost interpolation and clustering strategies, with the eventual goal of validating our approach over diverse terrain and factors such as varying lighting conditions.



\section*{ACKNOWLEDGEMENTS}

The authors would like to thank Jean-Pierre de la Croix, Federico Rossi, Changrak Choi, and Issa Nesnas for insightful discussions and Daniel Neamati and Adyasha Mohanty for reviewing the drafts of this paper. 

\bibliographystyle{apalike}
\bibliography{references}

@inproceedings{ono2016data,
  title={{Data-Driven Surface Traversability Analysis for Mars 2020 Landing Site Selection}},
  author={Ono, Masahiro and Rothrock, Brandon and Almeida, Eduardo and Ansar, Adnan and Otero, Richard and Huertas, Andres and Heverly, Matthew},
  booktitle={2016 IEEE Aerospace Conference},
  pages={1--12},
  year={2016},
  organization={IEEE}
}

@inproceedings{tancik2023nerfstudio,
  title={{Nerfstudio: A Modular Framework for Neural Radiance Field Development}},
  author={Tancik, Matthew and Weber, Ethan and Ng, Evonne and Li, Ruilong and Yi, Brent and Wang, Terrance and Kristoffersen, Alexander and Austin, Jake and Salahi, Kamyar and Ahuja, Abhik and others},
  booktitle={ACM SIGGRAPH 2023 Conference Proceedings},
  pages={1--12},
  year={2023}
}

@article{suzuki20113d,
  title={{3D terrain reconstruction by small unmanned aerial vehicle using SIFT-based monocular SLAM}},
  author={Suzuki, Taro and Amano, Yoshiharu and Hashizume, Takumi and Suzuki, Shinji},
  journal={Journal of Robotics and Mechatronics},
  volume={23},
  number={2},
  pages={292--301},
  year={2011},
  publisher={Fuji Technology Press}
}

@inproceedings{schonberger2016structure,
  title={Structure-from-motion revisited},
  author={Schonberger, Johannes L and Frahm, Jan-Michael},
  booktitle={Proceedings of the IEEE conference on computer vision and pattern recognition},
  pages={4104--4113},
  year={2016}
}

@inproceedings{shah2018airsim,
  title={{AirSim: High-Fidelity Visual and Physical Simulation for Autonomous Vehicles}},
  author={Shah, Shital and Dey, Debadeepta and Lovett, Chris and Kapoor, Ashish},
  booktitle={Field and Service Robotics: Results of the 11th International Conference},
  pages={621--635},
  year={2018},
  organization={Springer}
}

@inproceedings{tzanetos2022ingenuity,
  title={{Ingenuity Mars Helicopter: From Technology Demonstration to Extraterrestrial Scout}},
  author={Tzanetos, Theodore and Aung, MiMi and Balaram, J and Grip, Havard Fjrer and Karras, Jaakko T and Canham, Timothy K and Kubiak, Gerik and Anderson, Joshua and Merewether, Gene and Starch, Michael and others},
  booktitle={2022 IEEE Aerospace Conference (AERO)},
  pages={01--19},
  year={2022},
  organization={IEEE}
}

@inproceedings{carsten2007global,
  title={{Global Path Planning on Board the Mars Exploration Rovers}},
  author={Carsten, Joseph and Rankin, Arturo and Ferguson, Dave and Stentz, Anthony},
  booktitle={2007 IEEE Aerospace Conference},
  pages={1--11},
  year={2007},
  organization={IEEE}
}

@inproceedings{goldberg2002stereo,
  title={{Stereo Vision and Rover Navigation Software for Planetary Exploration}},
  author={Goldberg, Steven B and Maimone, Mark W and Matthies, Larry},
  booktitle={Proceedings, IEEE aerospace conference},
  volume={5},
  pages={5--5},
  year={2002},
  organization={IEEE}
}

@inproceedings{biesiadecki2006mars,
  title={{The Mars Exploration Rover surface mobility flight software driving ambition}},
  author={Biesiadecki, Jeffrey J and Maimone, Mark W},
  booktitle={2006 IEEE Aerospace Conference},
  pages={15--pp},
  year={2006},
  organization={IEEE}
}

@inproceedings{abcouwer2021machine,
  title={{Machine Learning Based Path Planning for Improved Rover Navigation}},
  author={Abcouwer, Neil and Daftry, Shreyansh and del Sesto, Tyler and Toupet, Olivier and Ono, Masahiro and Venkatraman, Siddarth and Lanka, Ravi and Song, Jialin and Yue, Yisong},
  booktitle={2021 IEEE Aerospace Conference (50100)},
  pages={1--9},
  year={2021},
  organization={IEEE}
}

@article{daftry2022mlnav,
  title={{MLNav: Learning to Safely Navigate on Martian Terrains}},
  author={Daftry, Shreyansh and Abcouwer, Neil and Del Sesto, Tyler and Venkatraman, Siddarth and Song, Jialin and Igel, Lucas and Byon, Amos and Rosolia, Ugo and Yue, Yisong and Ono, Masahiro},
  journal={IEEE Robotics and Automation Letters},
  volume={7},
  number={2},
  pages={5461--5468},
  year={2022},
  publisher={IEEE}
}

@inproceedings{toupet2020ros,
  title={{A ROS-based Simulator for Testing the Enhanced Autonomous Navigation of the Mars 2020 Rover}},
  author={Toupet, Olivier and Del Sesto, Tyler and Ono, Masahiro and Myint, Steven and Vander Hook, Joshua and McHenry, Michael},
  booktitle={2020 IEEE Aerospace Conference},
  pages={1--11},
  year={2020},
  organization={IEEE}
}

@article{adamkiewicz_vision-only_2022,
	title = {Vision-{Only} {Robot} {Navigation} in a {Neural} {Radiance} {World}},
	volume = {7},
	issn = {2377-3766, 2377-3774},
	url = {https://ieeexplore.ieee.org/document/9712211/},
	doi = {10.1109/LRA.2022.3150497},
	number = {2},
	urldate = {2023-03-01},
	journal = {IEEE Robotics and Automation Letters},
	author = {Adamkiewicz, Michal and Chen, Timothy and Caccavale, Adam and Gardner, Rachel and Culbertson, Preston and Bohg, Jeannette and Schwager, Mac},
	month = apr,
	year = {2022},
	pages = {4606--4613},
}

@inproceedings{garbin_fastnerf_2021,
	address = {Montreal, QC, Canada},
	title = {{FastNeRF}: {High}-{Fidelity} {Neural} {Rendering} at {200FPS}},
	isbn = {978-1-66542-812-5},
	shorttitle = {{FastNeRF}},
	url = {https://ieeexplore.ieee.org/document/9710021/},
	doi = {10.1109/ICCV48922.2021.01408},
	urldate = {2023-03-01},
	booktitle = {2021 {IEEE}/{CVF} {International} {Conference} on {Computer} {Vision} ({ICCV})},
	publisher = {IEEE},
	author = {Garbin, Stephan J. and Kowalski, Marek and Johnson, Matthew and Shotton, Jamie and Valentin, Julien},
	month = oct,
	year = {2021},
	pages = {14326--14335},
}

@inproceedings{reiser_kilonerf_2021,
	address = {Montreal, QC, Canada},
	title = {{KiloNeRF}: {Speeding} up {Neural} {Radiance} {Fields} with {Thousands} of {Tiny} {MLPs}},
	isbn = {978-1-66542-812-5},
	shorttitle = {{KiloNeRF}},
	url = {https://ieeexplore.ieee.org/document/9710464/},
	doi = {10.1109/ICCV48922.2021.01407},
	urldate = {2023-03-01},
	booktitle = {2021 {IEEE}/{CVF} {International} {Conference} on {Computer} {Vision} ({ICCV})},
	publisher = {IEEE},
	author = {Reiser, Christian and Peng, Songyou and Liao, Yiyi and Geiger, Andreas},
	month = oct,
	year = {2021},
	pages = {14315--14325},
}

@inproceedings{giusti2023marf,
  title={{MaRF: Representing Mars as Neural Radiance Fields}},
  author={Giusti, Lorenzo and Garcia, Josue and Cozine, Steven and Suen, Darrick and Nguyen, Christina and Alimo, Ryan},
  booktitle={Computer Vision--ECCV 2022 Workshops: Tel Aviv, Israel, October 23--27, 2022, Proceedings, Part I},
  pages={53--65},
  year={2023},
  organization={Springer}
}

@article{mildenhall2021nerf,
  title={{NeRF: Representing Scenes as Neural Radiance Fields for View Synthesis}},
  author={Mildenhall, Ben and Srinivasan, Pratul P and Tancik, Matthew and Barron, Jonathan T and Ramamoorthi, Ravi and Ng, Ren},
  journal={Communications of the ACM},
  volume={65},
  number={1},
  pages={99--106},
  year={2021},
  publisher={ACM New York, NY, USA}
}

@inproceedings{rieber2022planning,
  title={{Planning for a Martian Road Trip - The Mars2020 Mobility Systems Design}},
  author={Rieber, Richard and McHenry, Michael and Twu, Philip and Stragier, Michael M},
  booktitle={2022 IEEE Aerospace Conference (AERO)},
  pages={01--18},
  year={2022},
  organization={IEEE}
}

@article{sherwood2001integrated,
  title={{An Integrated Planning and Scheduling Prototype for Automated Mars Rover Command Generation}},
  author={Sherwood, Rob and Mishkin, Andrew and Chien, Steve and Estlin, Tara and Backes, Paul and Cooper, Brian and Rabideau, Gregg and Engelhardt, Barbara},
  year={2001},
  publisher={Pasadena, CA: Jet Propulsion Laboratory, National Aeronautics and Space~…}
}

@article{chen2023catnips,
  title={{CATNIPS: Collision Avoidance Through Neural Implicit Probabilistic Scenes}},
  author={Chen, Timothy and Culbertson, Preston and Schwager, Mac},
  journal={arXiv preprint arXiv:2302.12931},
  year={2023}
}

@misc{karadayi, title={Moon Landscape in Environments - UE Marketplace}, url={https://www.unrealengine.com/marketplace/en-US/product/moon-landscape-01?sessionInvalidated=true}, journal={Unreal Engine}, author={Karadayi, Gokhan},  year={2018}}

@article{kirillov2023segment,
  title={{Segment Anything}},
  author={Kirillov, Alexander and Mintun, Eric and Ravi, Nikhila and Mao, Hanzi and Rolland, Chloe and Gustafson, Laura and Xiao, Tete and Whitehead, Spencer and Berg, Alexander C and Lo, Wan-Yen and others},
  journal={arXiv preprint arXiv:2304.02643},
  year={2023}
}

@inproceedings{ferguson2007field,
  title={{Field D*: An Interpolation-Based Path Planner and Replanner}},
  author={Ferguson, Dave and Stentz, Anthony},
  booktitle={Robotics Research: Results of the 12th International Symposium ISRR},
  pages={239--253},
  year={2007},
  organization={Springer}
}

@inproceedings{xiangli2022bungeenerf,
  title={{BungeeNeRF: Progressive Neural Radiance Field for Extreme Multi-scale Scene Rendering}},
  author={Xiangli, Yuanbo and Xu, Linning and Pan, Xingang and Zhao, Nanxuan and Rao, Anyi and Theobalt, Christian and Dai, Bo and Lin, Dahua},
  booktitle={European conference on computer vision},
  pages={106--122},
  year={2022},
  organization={Springer}
}

@incollection{vovk2013kernel,
  title={Kernel ridge regression},
  author={Vovk, Vladimir},
  booktitle={Empirical Inference: Festschrift in Honor of Vladimir N. Vapnik},
  pages={105--116},
  year={2013},
  publisher={Springer}
}

\end{document}